\documentclass[letterpaper, 10 pt, conference]{ieeeconf}  

\IEEEoverridecommandlockouts                              
\usepackage[linesnumbered,ruled]{algorithm2e}
\usepackage{float}
\usepackage{subfigure}
\usepackage{graphics} 
\usepackage{epsfig} 
\usepackage{mathrsfs}
\usepackage{times} 
\usepackage{amsmath} 
\usepackage{amssymb}  
\usepackage{color}
\usepackage[table]{xcolor}
\usepackage[T1]{fontenc}
\usepackage{bm}
\usepackage{cite}
\usepackage{hyperref}
\usepackage{multirow}
\hypersetup{
	colorlinks=true,
	linkcolor=blue,
	filecolor=blue,      
	urlcolor=blue,
	citecolor=cyan,
}

\title{\LARGE \bf
	Augmented Memory for Correlation Filters in Real-Time UAV Tracking } 
\author{Yiming Li$^{1}$, Changhong Fu$^{1,*}$, Fangqiang Ding$^{1}$, Ziyuan Huang$^{2}$, and Jia Pan$^{3}$
	\thanks{$^{1}$Yiming Li, Changhong Fu and Fangqiang Ding are with the School of Mechanical Engineering, Tongji University, 201804 Shanghai, China.
		{\tt\small changhongfu@tongji.edu.cn}}%
	\thanks{$^{2}$Ziyuan Huang is with the Advanced Robotics Centre, National University of Singapore, Singapore.
		{\tt\small ziyuan.huang@u.nus.edu}}%
	\thanks{$^{3}$Jia Pan is with the Computer Science Department, The University of Hong Kong, Hong Kong, China.
		{\tt\small panjia1983@gmail.com}}%
}

\begin{document}

\maketitle
\thispagestyle{empty}
\pagestyle{empty}

\begin{abstract}
The outstanding computational efficiency of discriminative correlation filter (DCF) fades away with various complicated improvements. Previous appearances are also gradually forgotten due to the exponential decay of historical views in traditional appearance updating scheme of DCF framework, reducing the model's robustness. In this work, a novel tracker based on DCF framework is proposed to augment memory of previously appeared views while running at real-time speed. Several historical views and the current view are simultaneously introduced in training to allow the tracker to adapt to new appearances as well as memorize previous ones. A novel rapid compressed context learning is proposed to increase the discriminative ability of the filter efficiently. Substantial experiments on UAVDT and UAV123 datasets have validated that the proposed tracker performs competitively against other 26 top DCF and deep-based trackers with over 40 FPS on CPU.
\end{abstract}
\section{introduction}
Unmanned arieal vehicle (UAV) object tracking has many applications such as target tracing~\cite{Chen2016IROS}, robot localization~\cite{Coltin2016IROS}, mid-air tracking~\cite{Fu2014ICRA} and aerial cinematography~\cite{Bonatti2019IROS}. It aims to locate the object in the following frames given the initial location, in sometimes difficult situations such as fast motion, appearance variation (occlusion, illumination and viewpoint change, etc.), scale changes, and limited power capacity. 

In UAV tracking tasks, the speed has been a key issue besides its performance. It was because of its ability to track objects at hundreds of frames per second (FPS) that discriminative correlation filter (DCF) is widely applied to perform UAV tracking in the first place. Unfortunately, the pioneering works~\cite{Bolme2010CVPR,Henriques2012ECCV,Henriques2015TPAMI}, despite their incredible speed, have inferior tracking performances. Therefore, strategies like part-based methods~\cite{Liu2015CVPR,Fu2018ROBIO}, spatial punishment~\cite{Danelljan2015ICCV,Danelljan2016ECCV, Luke2017CVPR} and robust appearance representation~\cite{Danelljan2015ICCVW,Ma2015ICCV,Li2017ICCVW} are used to improve their precision and accuracy. However, speed of DCF is sacrificed in pursuit of better performances.

In order to adapt to appearance changes of tracked objects, an appearance model is maintained and updated at each frame for most DCF trackers. Due to its updating scheme, historical appearance decays exponentially with the number of subsequent frames. The appearance in latest 2 seconds in a 60 FPS video has a similar weight to all appearances before these 2 seconds in the model. This makes the trackers prone to forget objects' early appearances and focus on more recent ones. Therefore, when the tracker has a false result, when the object is occluded, or when it is out of the view, it is very likely that the tracker learns appearances of the background using this scheme, which will further lead to lost of object in the following frames.

\begin{figure}[t]
	\label{fig:scene}
	\includegraphics[width=\columnwidth]{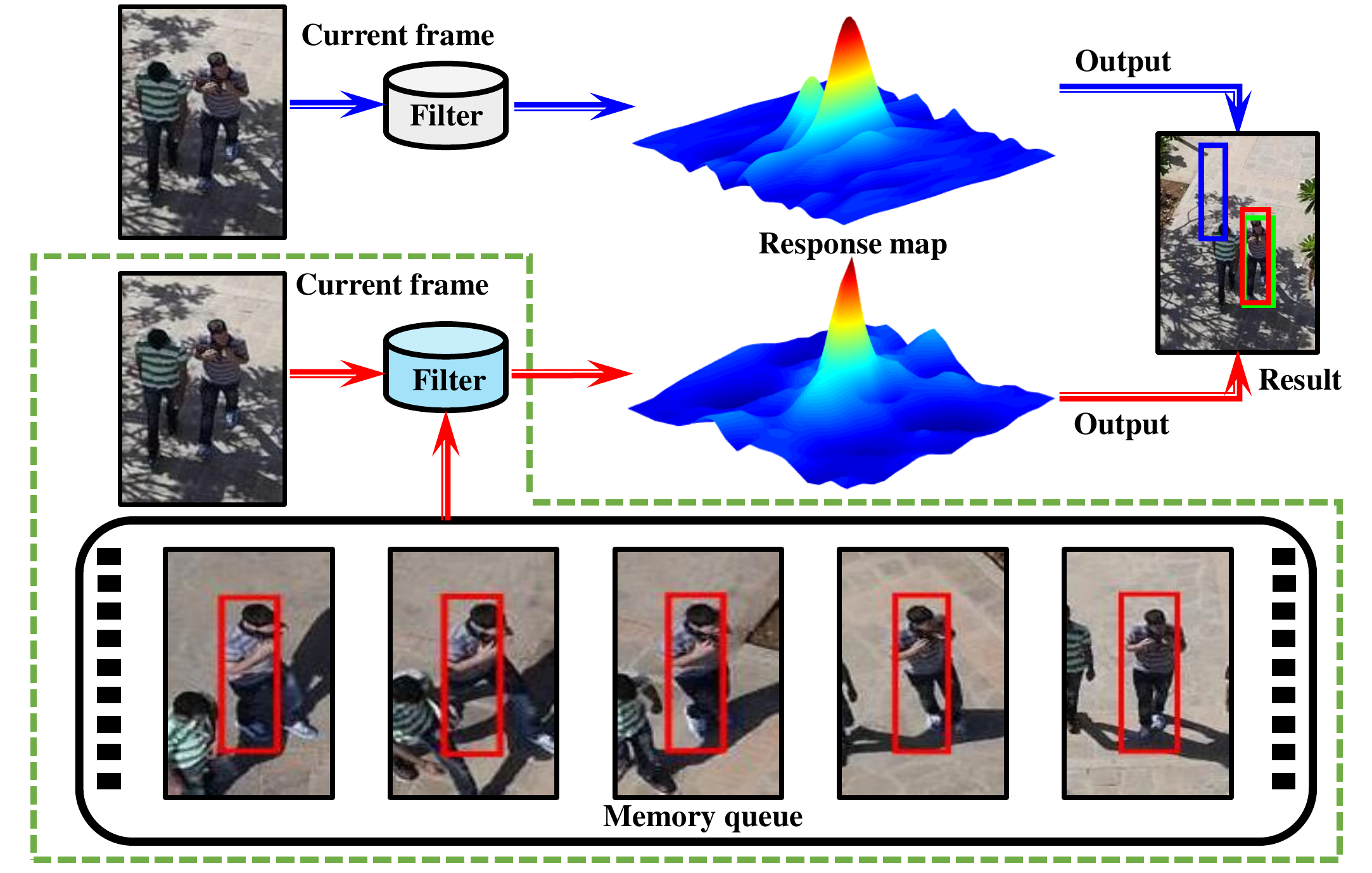}
	\caption{Comparison between traditional DCF with appearance models and DCF with the proposed augmented memory. Multiple historical views are selected and stored to be used in training so that it contains more historical appearance information. Traditional DCF uses an gradually decaying appearance model and is prone to drift when drastic appearance variation happens. Red and blue boxes denote the tracking results of our tracker and others respectively. Ground truth is displayed as green box.}
\end{figure}
Additionally, traditional DCF framework has a low discriminative power because of the lack of negative training samples. Spatial regularization~\cite{Danelljan2015ICCV,Danelljan2016ECCV, Luke2017CVPR} and target cropping~\cite{Galoogahi2015CVPR,Galoogahi2017ICCV,Huang2019ICCV,Fu2019IROS} were used to expand search region and extract background patches as negative samples. Introducing context and repressing response to it can also help discriminate objects from complex scenes~\cite{Mueller2017CVPR}. These methods all propose effective ways to solve the problem, but not efficient ones. 

This work proposes an augmented memory for correlation filters (AMCF) to perform efficient object tracking with strong discriminative power, which can easily be implemented on UAVs with only one CPU. Augmented memory is used to better memorize previous appearances of objects, with a novel application of image similarity criteria pHash~\cite{Kozat2001ICIP} to carefully select what to memorize. Views in the memory and current view are simultaneously used in training the correlation filter so that it has suitable responses to both previous and current object appearances. Compressed context learning is proposed to rapidly learn the background so that discriminative power is efficiently raised. AMCF is evaluated extensively on the authoritative UAVDT and UAV123 datasets. The results show its competitive performance on CPU at over 40 FPS compared with top trackers.
\section{related works}\label{sec:RELATEDWORK}
\subsection{Real-time tracking for UAV using DCF}
UAV object tracking has flourishing applications \cite{Fu2014ICUAS,2016YinTrans,Cheng2017IROS,Leira2017ICUAS,Bonatti2019IROS2}. Different from generic object tracking operating in videos shot by stationary cameras, tracking for UAV requires trackers to perform robustly even with the drastic motion of on-board camera. Motion of drone camera combined with that of objects makes precise and robust tracking difficult. Since in most cases, the movement of the drone is going to take largely depends on its perception result, tracking for UAV also requires a high processing speed of the tracker. 

Generative and discriminative methods have been applied to visual object tracking. One of the discriminative methods, DCF, has been widely adopted to perform the task. Although the original works~\cite{Bolme2010CVPR,Henriques2012ECCV,Henriques2015TPAMI} showed its exceeding performance in speed, they can hardly meet the requirement for accurate as well as robust tracking, and most current trackers have sacrificed their speed for better performances. A novel tracker that can balance the speed and performance is therefore called for.
\subsection{Model update scheme of DCF framework}
One significant difference between DCF and deep-learning based tracking methods is that DCF can track objects online without any pre-training on appearances of the objects. This is achieved because DCF framework usually maintains an appearance model that is updated on a frame-to-frame basis~\cite{Henriques2015TPAMI, Danelljan2015ICCV, Galoogahi2017ICCV, Li2018CVPR, Fu2019IROS, Huang2019ICCV}. To do that, mostly adopted scheme is that a new model in the new frame is composed of around 99\% of the previous model and around 1\% of the new appearance. This 1\% is treated as the learning rate. Some trackers use only new appearance as the new model~\cite{Li2018CVPR}. Unfortunately, the updating scheme causes a model decay, which means the appearance in early frames only takes up a small weight in this model. Therefore, when occlusion, out-of-view, or lost track of object happens, trackers tend to learn appearances of the background. This is not robust enough. 
\subsection{Negative samples and background noise in DCF}
The efficiency of DCF stems from its transformation of correlation operations into frequency domain. To do that, object image is cyclically shifted and extracted as samples implicitly. For traditional DCF framework, the search area is limited to prevent the correlation filter to learn too much from the background. However, only positive sample is essentially exploited in this manner. Several measures are taken to expand search region and feed background patches to training as negative samples~\cite{Galoogahi2017ICCV, Danelljan2015ICCV, Mueller2017CVPR, Fu2019IROS, Huang2019ICCV}. Typically, \cite{Danelljan2015ICCV} uses spatial punishment to suppress background learning, and \cite{Galoogahi2017ICCV} crops the target and background separately. \cite{Mueller2017CVPR} proposes to introduce context of the object and suppress the response to it. Despite their effectiveness, in order to be applied in UAV tracking, the efficiency of these methods is not sufficient.

\subsection{Tracking by deep learning} 
Deep learning is demonstrating its outstanding performance in various tasks. In DCF-based tracking, deep features are extracted by convolutional neural networks (CNN) in~\cite{Danelljan2015ICCVW,Ma2015ICCV,Li2017ICCVW} to further improve performance by strengthening object appearance representation. In addition to DCF-based tracking methods, end-to-end learning\cite{Valmadre2017CVPR}, deep reinforcement learning~\cite{Yun2017CVPR}, multi-domain network~\cite{Nam2016CVPR} and recurrent neural networks (RNNs)~\cite{Cui2016CVPR} directly use deep learning to perform tracking tasks. Despite their slightly superior performance, a high-end GPU is required for them to be trained and implemented. Even with that condition, most of them still run at low FPS. Therefore, deep-learning is not as suitable as DCF for aerial tracking tasks. 
\begin{figure*}[!t]
	\centering
	\includegraphics[width=0.95\textwidth]{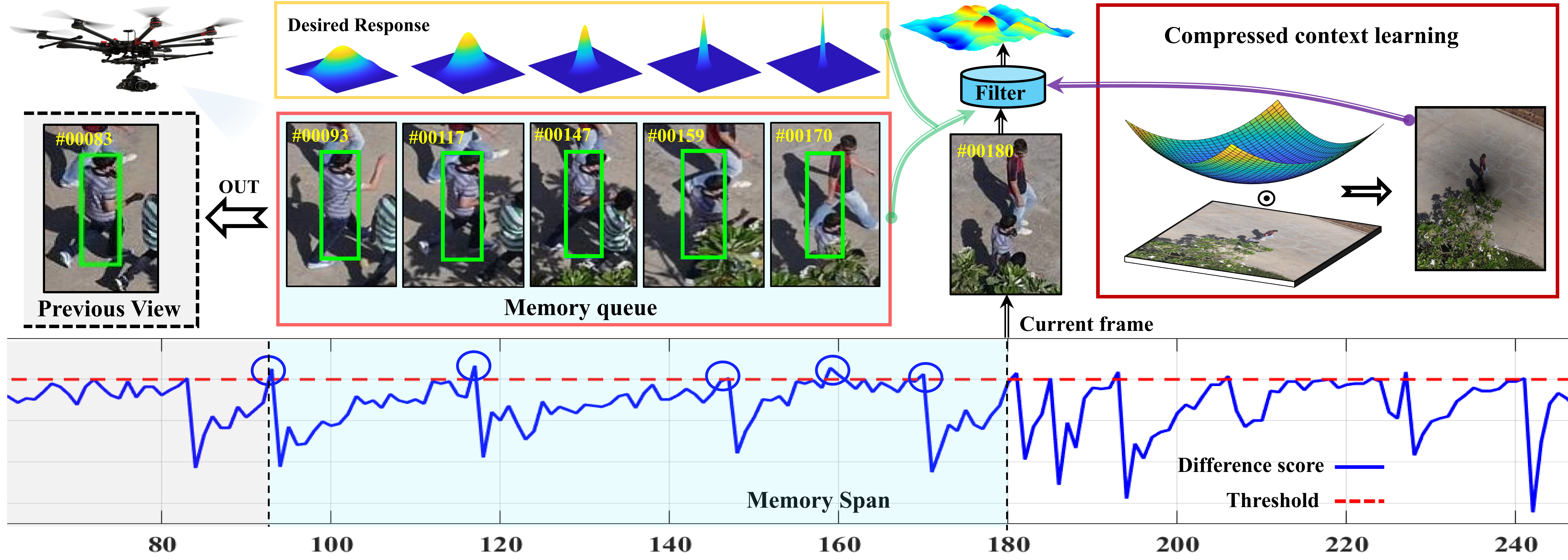}
	\caption{\textbf{Overall structure of AMCF.} Several historical views are stored in the memory queue and assigned with different desired responses according to its distance to the current frame. The selection of view is based on a difference score calculated by perceptual hashing algorithm. Along with views in the memory queue, context is also used in training. Context of the current frame is first compressed and given a weight before it is fed into training process. Tracking code and video can be seen respectively: \url{https://github.com/vision4robotics/AMCF-tracker} and \url{https://youtu.be/CGH5o2J1ohI}. }
	\label{fig:mainstructure}
\end{figure*}
\section{review of traditional dcf}\label{sec:DCF}
Learning traditional discriminative correlation filters~\cite{Galoogahi2017ICCV} is to optimize the following function:
	\begin{equation}\label{eq:dcfobjective}
	f(\mathbf{w}) = \Vert \mathbf { X }\mathbf { w }-\mathbf{y}\Vert _2^2 +\lambda \Vert \mathbf { w } \Vert _2^2 \ ,
	\end{equation}

\noindent where $\mathbf{X}$ is the sample matrix produced by circulating feature vector $\mathbf{x}\in\mathbb{R}^{M\times N}$, $\mathbf{w}$ is the trained filter and $\lambda$ is for regularization. It uses a model update scheme as follows:
\begin{equation}
\mathbf{m_t} = (1-\alpha )\mathbf{m_{t-1}} + \alpha \mathbf{x_t}\ ,
\end{equation}
\noindent where $\mathbf{m_t}$ and $\mathbf{ x_t }$ denotes the model and object appearance feature in the $t$-th frame respectively, $\alpha$ is the fixed learning rate.  Early appearances of the object are decaying exponentially. 
\section{augmented memory for dcf}\label{sec:METHOD}
In this section, learning augmented memory correlation filters for real-time UAV tracking is presented. The main structure of AMCF tracker is illustrated in Fig.~\ref{fig:mainstructure}, and the objective function is as follows:
\begin{equation}
\begin{aligned}
f_t(\mathbf{w})=&\Vert\mathbf{X}_{c}\mathbf{w}-\mathbf{y}_{c}\Vert_2^2+\lambda_2\sum_{k=0}^{K}\Vert\mathbf{X}_{k}\mathbf{w}-\mathbf{y}_{k}\Vert_2^2\\
&+\lambda_3\Vert (\mathbf{H}\odot \mathbf{X}_b) \mathbf{w}\Vert_2^2+\lambda_1\Vert \mathbf{w}\Vert_2^2 
\end{aligned} \ ,
\end{equation}

\noindent where $\mathbf{X}_c$, $\mathbf{X}_k$ and $\mathbf{X}_b$ presents the sample matrix generated by circulating feature maps $\mathbf{x}_c$, $\mathbf{x}_k$ and $\mathbf{x}_b$. $\mathbf{x}_{c}$ represents the extracted patch in current frame, $\mathbf{x}_0$ is the training patch sampled at the first frame, $\mathbf{x}_{k} \ (k\in \{1,2, ... ,$K$\}, K<<M)$ is the $k$-th view from the memory queue introduced in \ref{subsec:Selection}, where $M$ is the length of the sequence. $\mathbf{y}_{c}$ and $\mathbf{y}_{k}$ are distinct desired response of current frame and the ones in the memory (explained in \ref{subsec:Response}). $\mathbf{x}_b$ denotes compressed context patch, $\mathbf{H}$ refers to a suppression matrix generated by circulating $\mathbf{h} \in \mathbb{R}^{M\times N} $ and its function is to remove the object from the compressed context (explained in \ref{subsec:Context}). $\lambda_k \ (k=1,2,3)$ are adjustable parameters that determines the importance of corresponding patch.
\subsection{Augmented memory}\label{subsec:Selection}
\subsubsection{Memory queue}
Basically, a first-in first-out queue is maintained with a total length of $K$ to store $K$ historical views, so that they can be exploited by the training of a correlation filter in each frame. Before the memory queue is full, i.e., $L<K$ (with $L$ being the number of views currently stored), only $L$ views are fed in the training process. Otherwise, all $K$ historical views are used to train the correlation filter. For efficiency reasons, the value of $K$ is significantly smaller than the total number of the sequence. 
\subsubsection{View selection}
Since the number of views that can be stored is limited, it is important that different views contain different appearance details. Therefore, only when the appearance of object is significantly different than the last selected view is this appearance allowed into the memory queue. Perceptual hashing algorithm (PHA)~\cite{Kozat2001ICIP} is adopted to determine the level of difference between two appearances. Specifically, the gray images are firstly transformed to frequency domain by discrete cosine transform (DCT). Only the low frequency region with high energy density is retained, denoted as $\mathbf{B} \in \mathbb{R}^{S\times S}$. Every element $b_{ij}$ ($i$ and $j$ denote the index of the element) in $\mathbf{B}$ is then compared with the average value of $\mathbf{B}$ to generate respective element of $p_{ij}$ in image's hashing matrix $\mathbf{P}\in \mathbb{R}^{S\times S}$:
\begin{equation}
if \ b_{ij}> \sum_{i=1}^{S}\sum_{j=1}^{S} b_{ij}/S^2 \ then \ p_{ij}=1; \ else \  p_{ij}=0.
\end{equation}

Difference score of the last view and the current view is calculated using respective hashing matrices $\mathbf{P}^l$ and $\mathbf{P}^c$:
\begin{equation}
score=\frac{\sum_{i=1}^{S}\sum_{j=1}^{S}(p_{ij}^c\oplus p_{ij}^l)}{S^2} \ ,
\end{equation}
where $p_{ij}^c$ and $p_{ij}^l$ respectively denote the element of hashing matrix $\mathbf{P}^c$ and $\mathbf{P}^l$. And $\oplus$ is the XOR operator.
If the score is more than threshold $\tau$, two appearances are considered different and current one is selected into the memory queue.
\subsection{Different desired responses}\label{subsec:Response}
Earlier selected views generally have a lower similarity to the current appearance than the later selected ones. Therefore, different desired responses are assigned to different views in the memory queue. By altering the maximum and the variance of Gaussian function, lower maximum and larger variance of desired responses are generated for early views:
\begin{equation}
\begin{aligned}
&\mathbf{y}_{K}(\sigma_{K})=\nu \mathbf{y}_{c}(\mu\sigma_c)\\
& \mathbf{y}_{k}(\sigma_{k})=\nu \mathbf{y}_{k+1}(\mu\sigma_{k+1}) \ (k=1,2,...,K-1) 
\end{aligned}\ ,
\end{equation}
\noindent  where $\mathbf{y}_c(\mu\sigma_c)$ denotes the desired response of the current frame with the maximum of $max(\mathbf{y}_c(\mu\sigma_c))$ and the variance of $\sigma_c$. The subscript of $\mathbf{y}$ represents the index value of response target in memory queue (lower values corresponds to earlier views). Parameters $\nu<1$ and $\mu>1$ make sure that maximum is decreasing and variance is increasing with the index decreasing. For the first frame of the sequence, the desired response is calculated as follows:
\begin{equation}
\mathbf{y}_0(\sigma_1)=\phi \mathbf{y}_c(\varphi\sigma_c) \ ,
\end{equation}
\noindent where $\phi<1$ and $\varphi>1$ are used to adjust the maximum and variance of the Gaussian distribution of the desired response.
\subsection{Compressed context learning}\label{subsec:Context}
In order to increase discriminative power of DCF, compressed context learning is proposed. Unlike traditional context learning, the enlarged search region is compressed to the size of the correlation filter. Then the pixel value where the object is located is lowered by applying a quadratic suppressing function, so as to remove the object from this patch. This compressed context is assigned zero response so that the response to the surrounding area of the object can be minimized and discriminative power can thus be enhanced.
\subsection{Learning and detection in AMCF}\label{subsec:solution}
\subsubsection{Learning process}
The optimization result of $f_t(\mathbf{w})$ for the $d$-th $(d\in \{1,...,D\})$ channel is calculated as follows:

\begin{equation}\label{eq:8}
\hat{\mathbf{w}}^d=\frac{\hat{\mathbf{M}}_{c}^d+\lambda_2\sum_{k=0}^{K}\hat{\mathbf{M}}_k^d}{\sum_{d=1}^{D}(\hat{\mathbf{A}}_{c}^d+\lambda_2\sum_{k=0}^{K}\hat{\mathbf{A}}_k^d+\lambda_3\hat{\mathbf{E}}^d)+\lambda_1} \ ,
\end{equation}
\noindent where $\hat{\mathbf{M}}=\hat{\mathbf{x}}^*\odot \hat{\mathbf{y}}$, $\hat{\mathbf{A}}=\hat{\mathbf{x}}^*\odot \hat{\mathbf{x}}$ and  $\hat{\mathbf{E}}=\hat{\mathbf{m}}^*_b\odot \hat{\mathbf{m}}_b$. 
$\mathbf{m}_b=\mathbf{h}\odot \mathbf{x}_b$. $\mathbf{x}^*$ denotes the complex-conjugate of $\mathbf{x}$ and the operator $\odot$ stands for the element-wise product. Hat mark is the discrete Fourier transform (DFT) value. 

A learning rate $\gamma$ is used to update the numerator $\hat{\mathbf{N}}_t^d$ and the denominator $\hat{\mathbf{D}}_t^d$ of the filter $\hat{\mathbf{w}}_t^d$ in the $t$-th frame:
\begin{small}
	\begin{equation}\label{eq:9}
	\begin{aligned}
	&\hat{\mathbf{w}}^d_t=\frac{\hat{\mathbf{N}}_t^d}{\sum_{d=1}^{D}\hat{\mathbf{D}}_t^d+\lambda_1}\\
	&\hat{\mathbf{N}}_t^d=(1-\gamma)\hat{\mathbf{N}}_{t-1}^d+\gamma\ ( \hat{\mathbf{M}}_{tc}^d+\lambda_2\sum_{k=0}^{K}\hat{\mathbf{M}}_{tk}^d)\\
	&\hat{\mathbf{D}}_t^d=(1-\gamma)\hat{\mathbf{D}}_{t-1}^d+\gamma\ (
	\hat{\mathbf{A}}_{tc}^d+\lambda_2\sum_{k=0}^{K}\hat{\mathbf{A}}_{tk}^d+\lambda_3\hat{\mathbf{E}}_{t}^d)
	\end{aligned}\ .
	\end{equation}
\end{small}

In order to make sure reliable channels contribute more to the final result, a channel weight $\mathbf{C}=\{c^d\}(d\in\{ 1,...,D\})$ is assigned to each channel and updated as follows:
\begin{equation}\label{eq:10}
	c^d_t=(1-\eta) c^d_{t-1}+\eta \frac{max(\hat{\mathbf{w}}^{d*}_t \odot \hat{\mathbf{x}}^d_{tc})}{\sum_{d=1}^{D}max(\hat{\mathbf{w}}^{d*}_t \odot \hat{\mathbf{x}}^d_{tc})}\ .
\end{equation}
\subsubsection{Detection in AMCF}
In detection phase, the following formula is used to generate the final response map $\mathbf{R}_t$ and update the position by searching the maximum value:
\begin{equation}
	\mathbf{R}_t=\mathscr{F}^{-1}(\sum_{d=1}^{D}{c}^d_t\hat{\mathbf{w}}^{d*}_{t-1} \odot \hat{\mathbf{z}}^d_t),
\end{equation}
\noindent where $\hat{\mathbf{w}}^{d*}$ and $\hat{\mathbf{z}}^d_t$ are respectively learned filter and current feature of search region in frequency domain, and $\mathscr{F}^{-1}$ denotes the inverse discrete Fourier transformation (IDFT).

\begin{algorithm}[h]
	\caption{AMCF tracker}
	\KwIn{	Groundtruth  in the first frame\\
		Subsequent frames
	}	
	\KwOut{	Predicted position of target in $t>1$ frame}
	\label{alg:KAOTtrackerflow}
	\eIf{$t=1$}{
		Extract $\mathbf{x}_{f}$ and $\mathbf{x}_b$ centered at the groundtruth\\
		Use Eq.~(\ref{eq:8}) to initialize the filters $\mathbf{w}_1$\\
		Initialize channel weight model $\{c^d\}=1/D$ 
	}
	{
		Extract $\mathbf{z}_t$ centered at location on frame $t-1$\\
		Use Eq.~(\ref{eq:10}) to generate the response map\\
		Find the peak position of map and output\\
		Extract $\mathbf{x}_{t}$ and $\mathbf{x}_b$ centered at location on frame $t$\\
		Calculate the $score$  between $\mathbf{P}_L$ and $\mathbf{P}_t$\\
		\If{score>$\tau$}
		{
				Update FIFO memory queue
		}
		Use Eq.~(\ref{eq:8}) to update the filters $\mathbf{w}_t$ \\
		Use Eq.~(\ref{eq:9}) to update channel weight $\mathbf{C}_t$\\
	}	
\end{algorithm} 
\section{EXPERIMENTS}\label{sec:EXPERIMENT}
In this section, the presented AMCF tracker is comprehensively evaluated on two difficult datasets, i.e., UAVDT~\cite{Du2018ECCV} and UAV123~\cite{Mueller2016ECCV}, with 173 image sequences covering over 140,000 frames captured by UAV in various challenging scenarios. It is noted that the videos from both datasets are recorded at 30 FPS. 11 real-time trackers (CPU based) are used to compare with AMCF, i.e., ECO\_HC~\cite{Danelljan2017CVPR}, STRCF~\cite{Li2018CVPR}, MCCT\_H~\cite{Wang2018CVPR}, STAPLE\_CA~\cite{Mueller2017CVPR}, BACF~\cite{Galoogahi2017ICCV}, DSST~\cite{Danelljan2014BMVA}, fDSST~\cite{Danelljan2017TPAMI}, STAPLE~\cite{Bertinetto2016TPAMI}, KCC~\cite{wang2018kernel}, KCF~\cite{Henriques2015TPAMI}, DCF~\cite{Henriques2015TPAMI}. Furthermore, 15 deep-based trackers are compared with AMCF to further demonstrate its performance, i.e., ASRCF~\cite{Kenan2019CVPR}, ECO~\cite{Danelljan2017CVPR}, C-COT~\cite{Danelljan2016ECCV}, MCCT~\cite{Wang2018CVPR}, DeepSRTCF~\cite{Li2018CVPR}, ADNet~\cite{Yun2017CVPR}, CFNet~\cite{Valmadre2017CVPR},  MCPF~\cite{Zhang2017CVPR}, IBCCF~\cite{Li2017ICCVW}, CF2~\cite{Ma2015ICCV}, CREST~\cite{Song2017ICCV}, HDT~\cite{Qi2016CVPR}, FCNT~\cite{Wang2015ICCV}, PTAV~\cite{Fan2017ICCV}, TADT~\cite{Xin2019CVPR}. The evaluation criteria are strictly according to the protocol in two benchmarks~\cite{Du2018ECCV,Mueller2016ECCV}.
\begin{table}[!b]
	\setlength{\tabcolsep}{2mm}
	\centering
	\caption{Effectiveness study of AMCF on UAV123 and UAVDT. Module name displayed in abbreviations are CW (channel weight), AM (augmented memory) and CC (compressed context). }
	\begin{tabular}{l|c|c|c|c|c|c}
		\hline
		Dateset&\multicolumn{3}{c|}{UAV123}&\multicolumn{3}{c}{UAVDT}\\
		\hline
		Evaluation&PREC.&AUC&FPS&PREC.&AUC&FPS\\
		\hline
		AMCF&69.5&49.3&38.1&70.1 & 44.5&46.7 \\
		Baseline+CW+CC&68.5&48.3&42.6&67.3&44.0&53.1\\
		Baseline+CC+AM&67.4&47.8&42.6 &69.0&44.4&52.1\\
		Baseline+CW&66.8&46.8& 50.4 &67.9&44.0&65.5 \\
		Baseline&65.5&46.8& 59.1 &66.4&42.9&73.5 \\
		\hline
	\end{tabular}%
	\label{tab:effectivenssstudy}%
\end{table}%
\begin{figure*}[!t]
	\centering
	\includegraphics[width=0.49\linewidth]{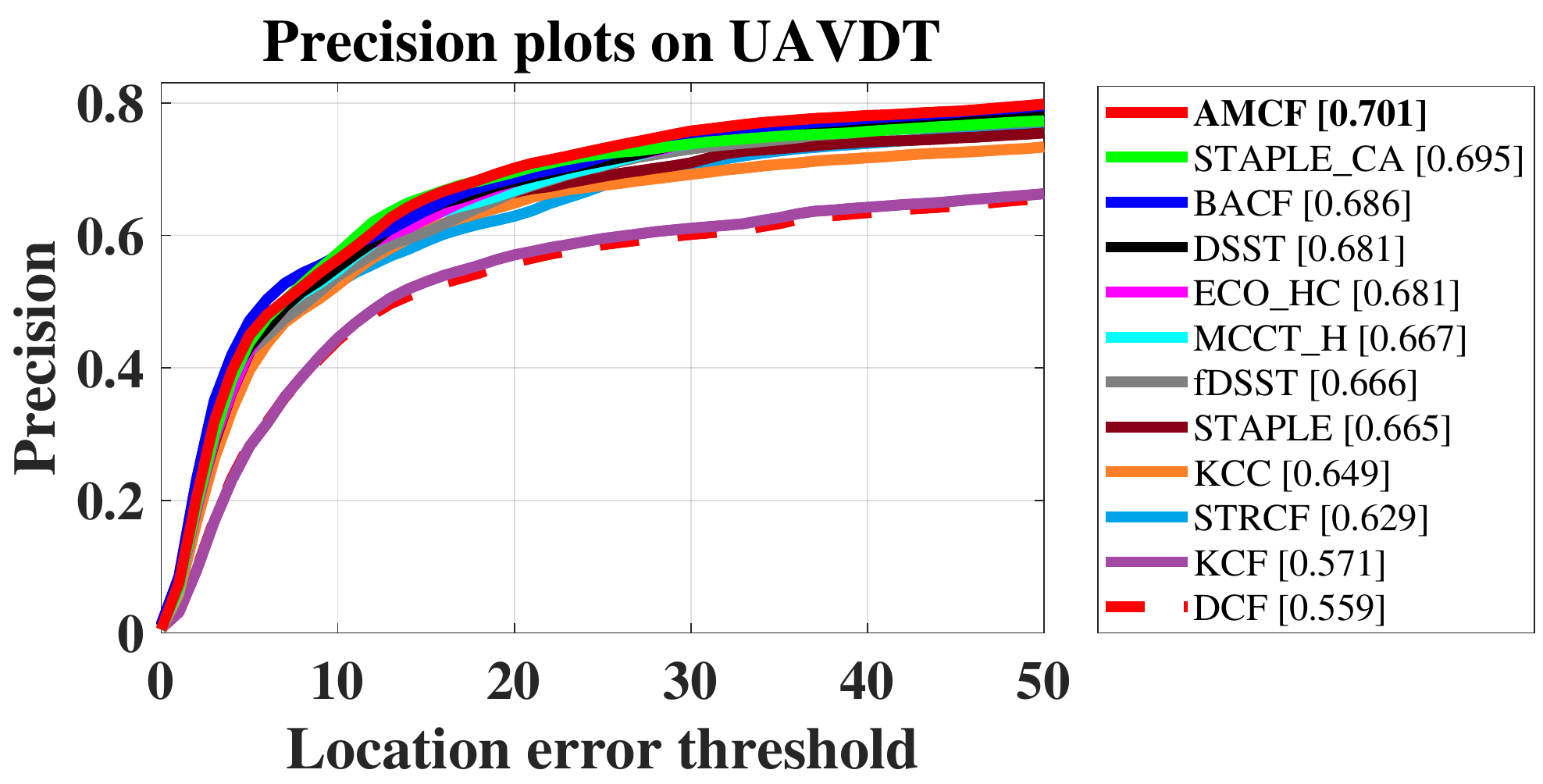}
	\includegraphics[width=0.49\linewidth]{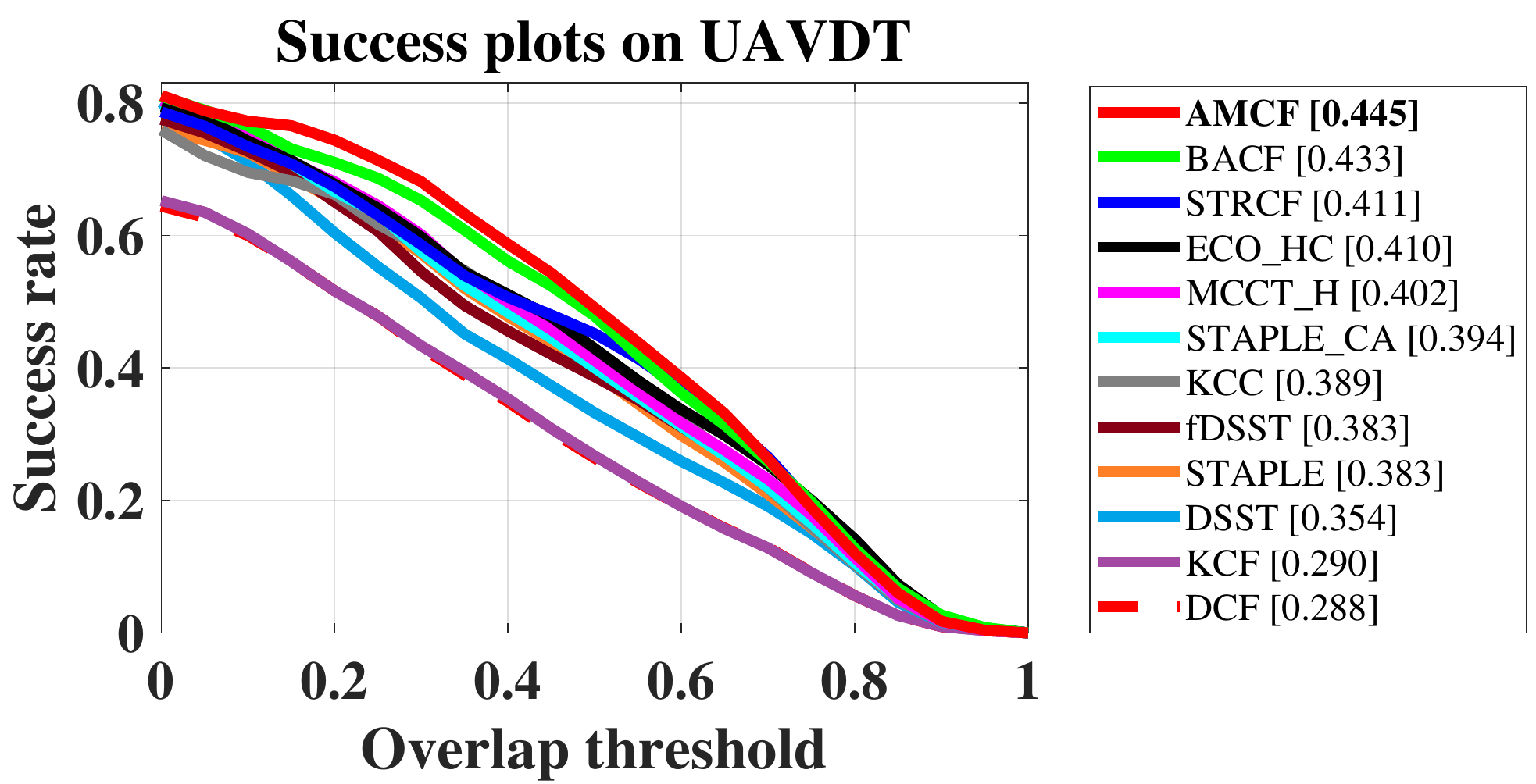}
	\\
	\includegraphics[width=0.49\linewidth]{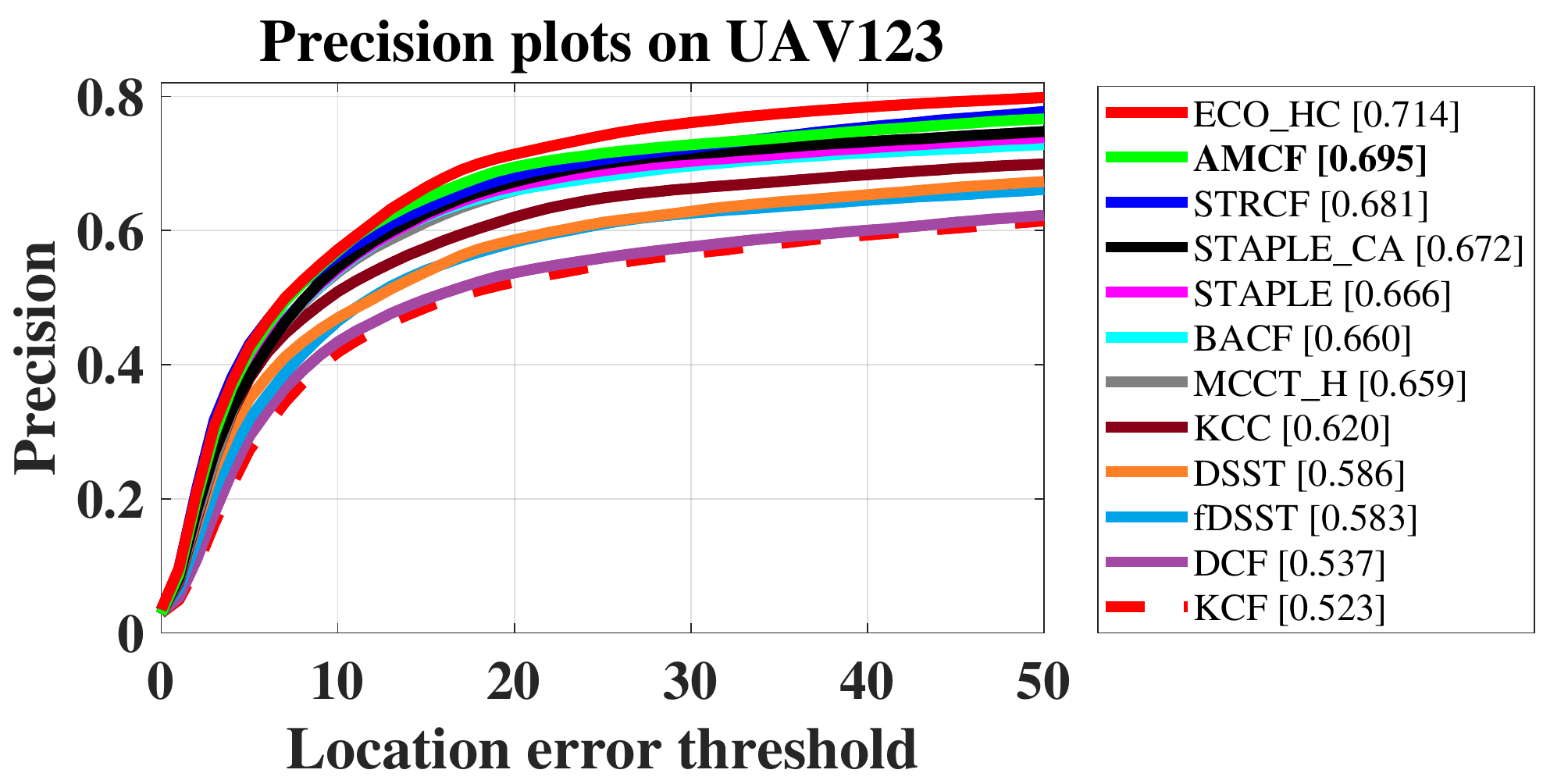}
	\includegraphics[width=0.49\linewidth]{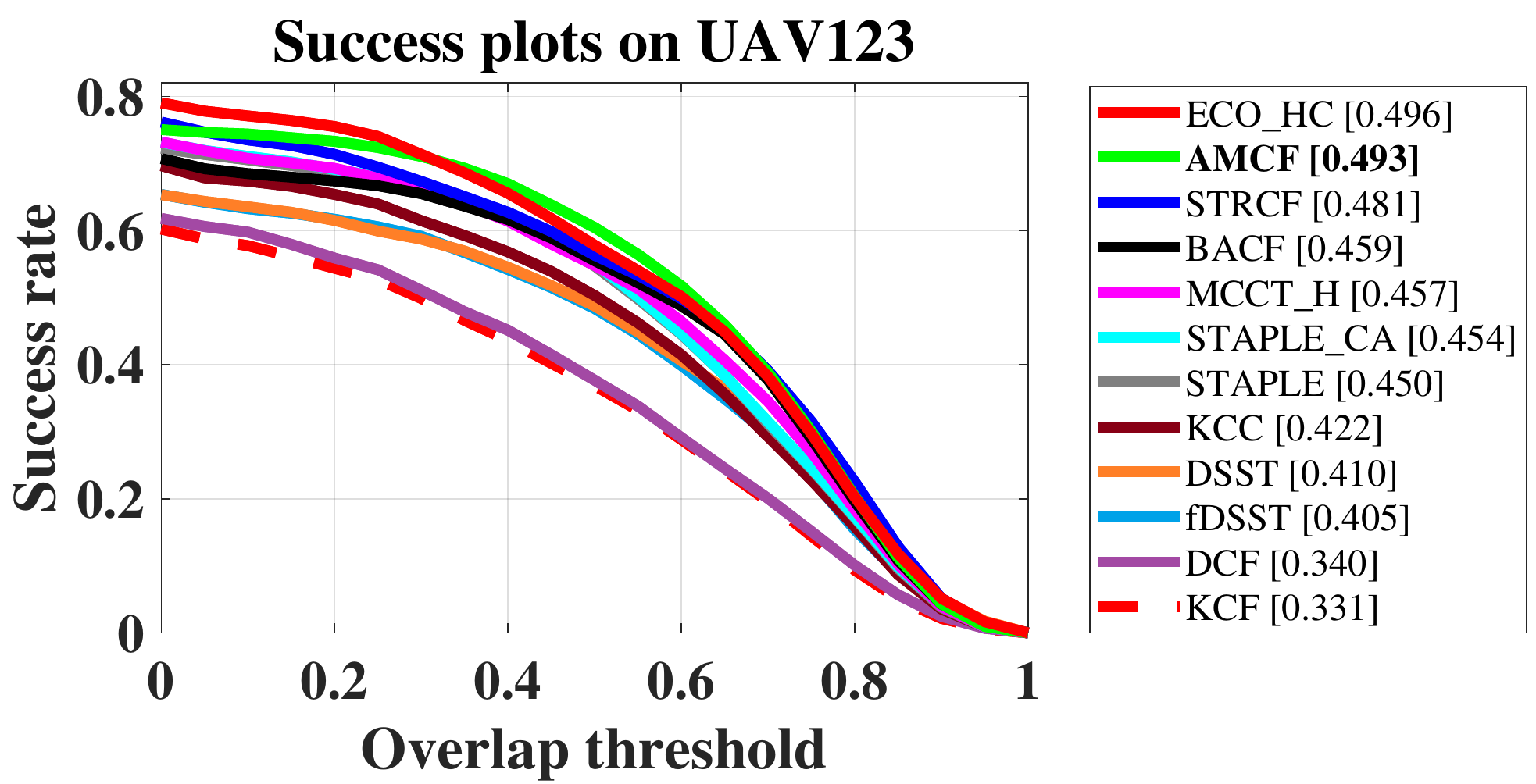}
	\caption{\textbf{Overall performance evaluation.} Precision and success plots of our tracker and other ten top real-time trackers on UAVDT and UAV123 datasets.}
	\label{fig:overall}
\end{figure*}
\begin{table*}[!t]
	\footnotesize
	\setlength{\tabcolsep}{2mm}
	\centering
	\caption{Average frame per second (FPS) and average precision as well as AUC of top real-time trackers on 173 image sequences. \textcolor[rgb]{ 1,  0,  0}{\textbf{Red}}, \textcolor[rgb]{ 0,  1,  0}{\textbf{green}} and \textcolor[rgb]{ 0,  0,  1}{\textbf{blue}} fonts indicate the first, second and third place, respectively. All results are obtained solely on CPU.}
	\begin{tabular}{l c c c c c c c c c c c c}
		\hline
		\multirow{2}{*}{\textbf{Tracker}}&\multirow{2}{*}{\textbf{AMCF}}&{ECO\_HC}&{MCCT\_H}&{STAPLE\_CA}&{KCC}&{BACF}&{STRCF}&{STAPLE}&{DSST}&{fDSST}&{KCF}&{DCF} \\ 
		&&\cite{Danelljan2017CVPR}&\cite{Wang2018CVPR}&\cite{Mueller2017CVPR}&\cite{wang2018kernel}&{\cite{Galoogahi2017ICCV}}&{\cite{Li2018CVPR}}&{\cite{Bertinetto2016TPAMI}}&{\cite{Danelljan2014BMVA}}&{\cite{Danelljan2017CVPR}}&{\cite{Henriques2015TPAMI}}&{\cite{Henriques2015TPAMI}} \\ 
		\hline
		\textbf{FPS} & {42.4} & {77.4} & {59.9} & {60.2} &{48.9}&{58.8}&{29.8}&{84.5}&{124.1}&\textbf{\textcolor[rgb]{0 0 1}{186.1}}&\textbf{\textcolor[rgb]{0 1 0}{795.7}}&\textbf{\textcolor[rgb]{1  0  0}{1120.7}}\\ 
		\textbf{Precision} & \textbf{\textcolor[rgb]{1  0  0}{69.8}} & \textbf{\textcolor[rgb]{1  0  0}{69.8}} & {66.1} & \textbf{\textcolor[rgb]{0 1 0}{67.7}} &{62.6}&{66.6}&\textbf{\textcolor[rgb]{0 0 1}{67.0}}&{66.6}&{60.7}&{60.1}&{53.3}&{54.2}\\ 
		\textbf{AUC} & \textbf{\textcolor[rgb]{1  0  0}{46.9}} & \textbf{\textcolor[rgb]{0  1  0}{45.3}} & {43.0} & {42.4} & {40.6} & \textbf{\textcolor[rgb]{0  0  1}{44.6}} & \textbf{\textcolor[rgb]{0  0  1}{44.6}} & {41.7} & {38.2} & {39.4} & {31.1} & {34.4}\\
		\hline
	\end{tabular}%
	\label{tab:fps}%
\end{table*}%
\subsection{Implementation details}
\label{subsec:EvaCri}
All the experiments of all trackers compared as well as AMCF are conducted on a computer with an CPU of i7-8700K (3.7GHz), 48GB RAM and NVIDIA GTX 2080. All trackers are implemented in MATLAB R2018a platform, and their original codes without modification are used for comparison. Memory length $K=5$, and the threshold for memory view selection is set to $\tau = 0.5$.

\subsection{Quantitative study}
\subsubsection{Effectiveness study}
AMCF tracker is firstly compared with itself with different modules enabled. The effectiveness evaluation result can be seen in Table~\ref{tab:effectivenssstudy}. With each module (channel weight CW, augmented memory AM, and compressed context CC) added to the baseline, the performance is steadily being improved.

\subsubsection{Overall performance}
In comparison with other top real-time trackers, AMCF has shown a superiority in terms of precision and accuracy on both benchmarks. Fig.~\ref{fig:overall} shows separate performance evaluation of AMCF on two benchmarks. AMCF has achieved satisfactory performances on both benchmarks. Specifically, AMCF performs the best on UAVDT, with improvement of 0.6\% and 1.2\% on precision and AUC score respectively. On UAV123, AMCF has achieved the second best performance. Since the object size in UAV123 is generally much larger than that in UAVDT because of the flying height of UAVs, many trackers with good performance on one benchmark can rank low on the other. One typical example is ECO\_HC. It ranks first on UAV123 but only come out at fifth place on UAVDT. AMCF, on the other hand, has a better generalization ability compared to most trackers. Overall evaluation for both benchmarks combined can be seen in Table~\ref{tab:fps}. AMCF has a slightly better overall performance than ECO\_HC, ranking the first place. But ECO\_HC has a relatively large variance. Therefore, it can be concluded that in terms of overall performance, AMCF performs favorably against other top real-time trackers. In terms of speed, AMCF, soly running on CPU, can also meet the requirement of real-time tracking (>30 FPS on a image sequence captured at 30 FPS).
\begin{figure*}[t]
	\centering	
	\includegraphics[width=0.245\linewidth]{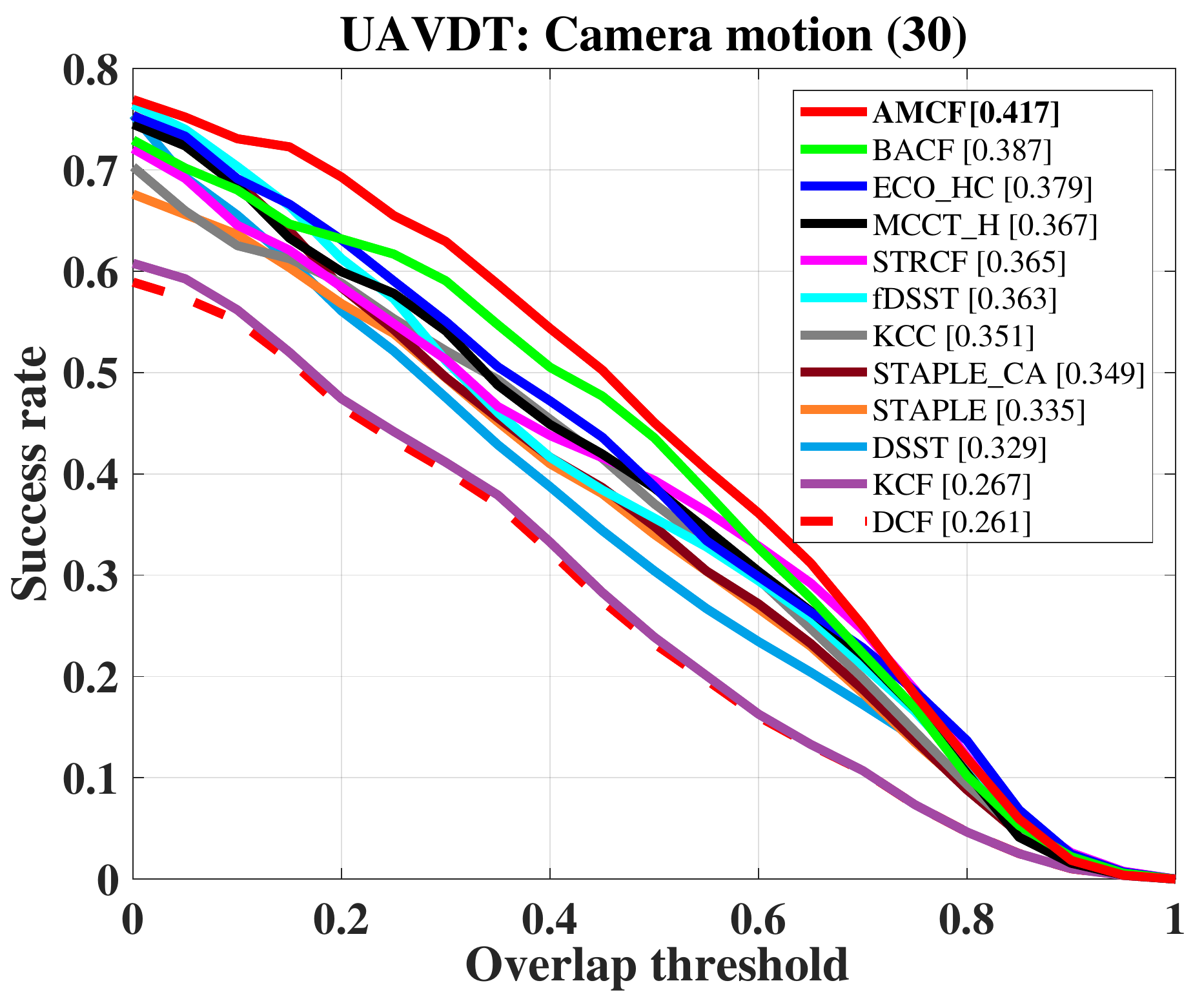}
	\includegraphics[width=0.245\linewidth]{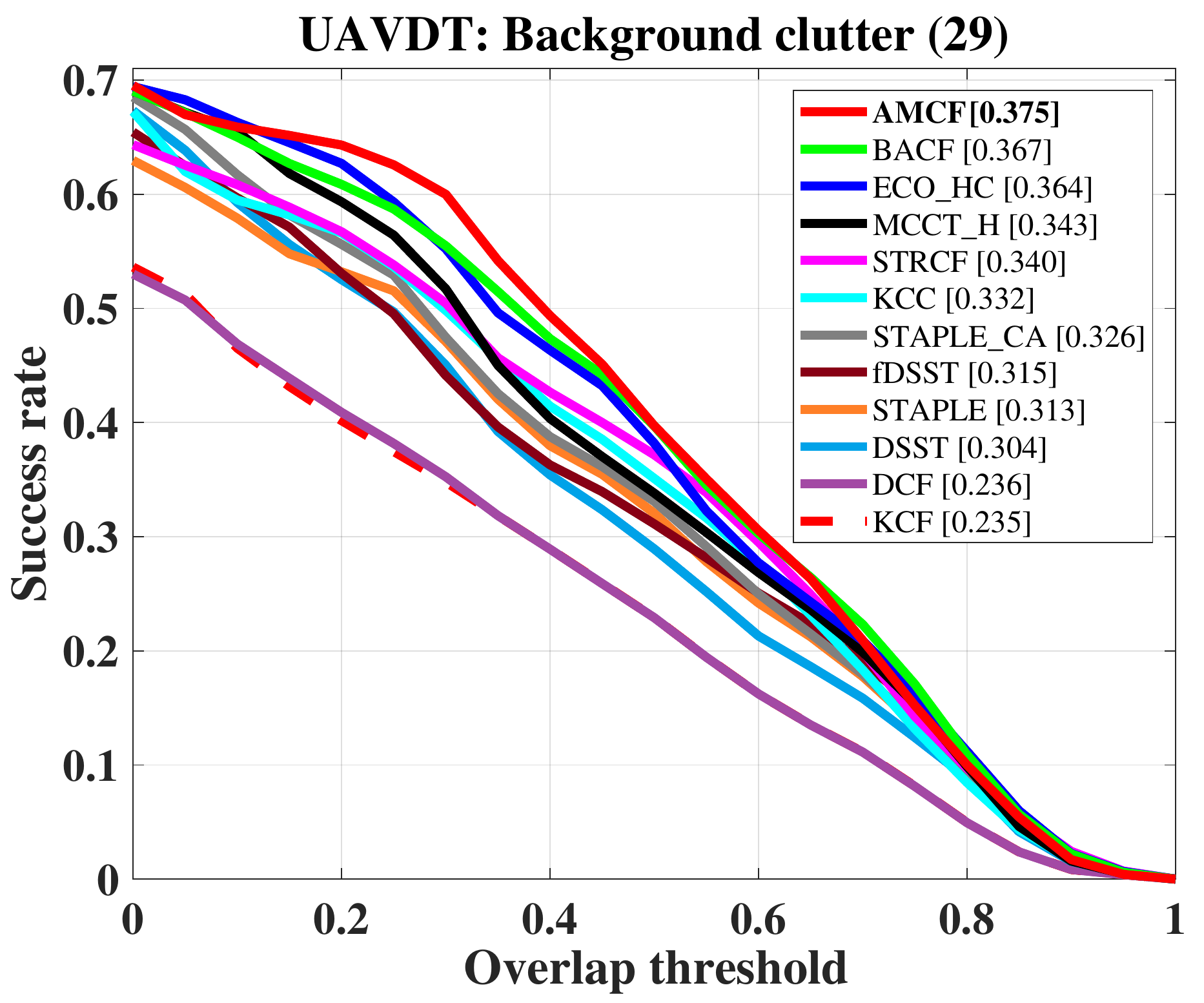}
	\includegraphics[width=0.245\linewidth]{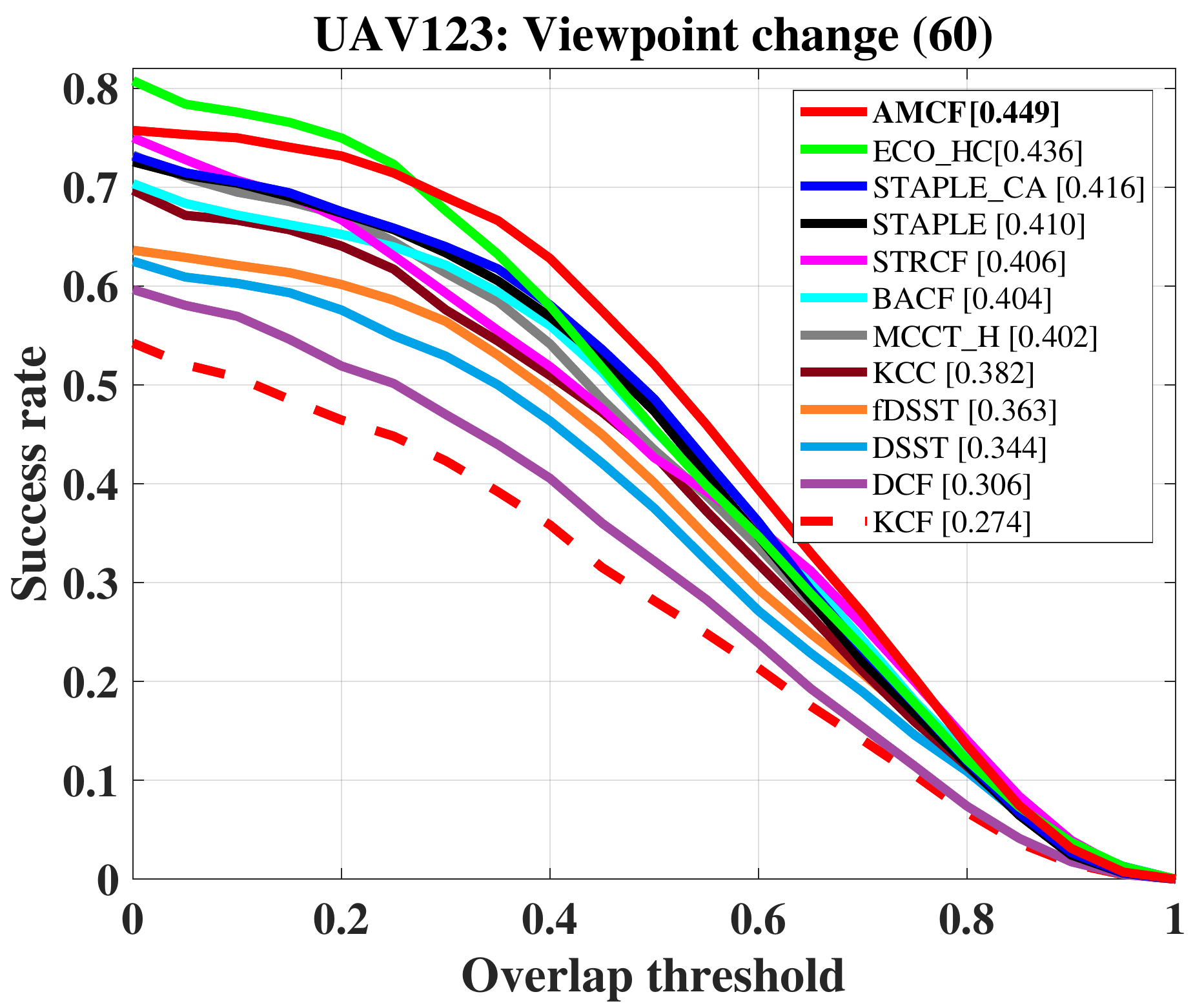}
	\includegraphics[width=0.245\linewidth]{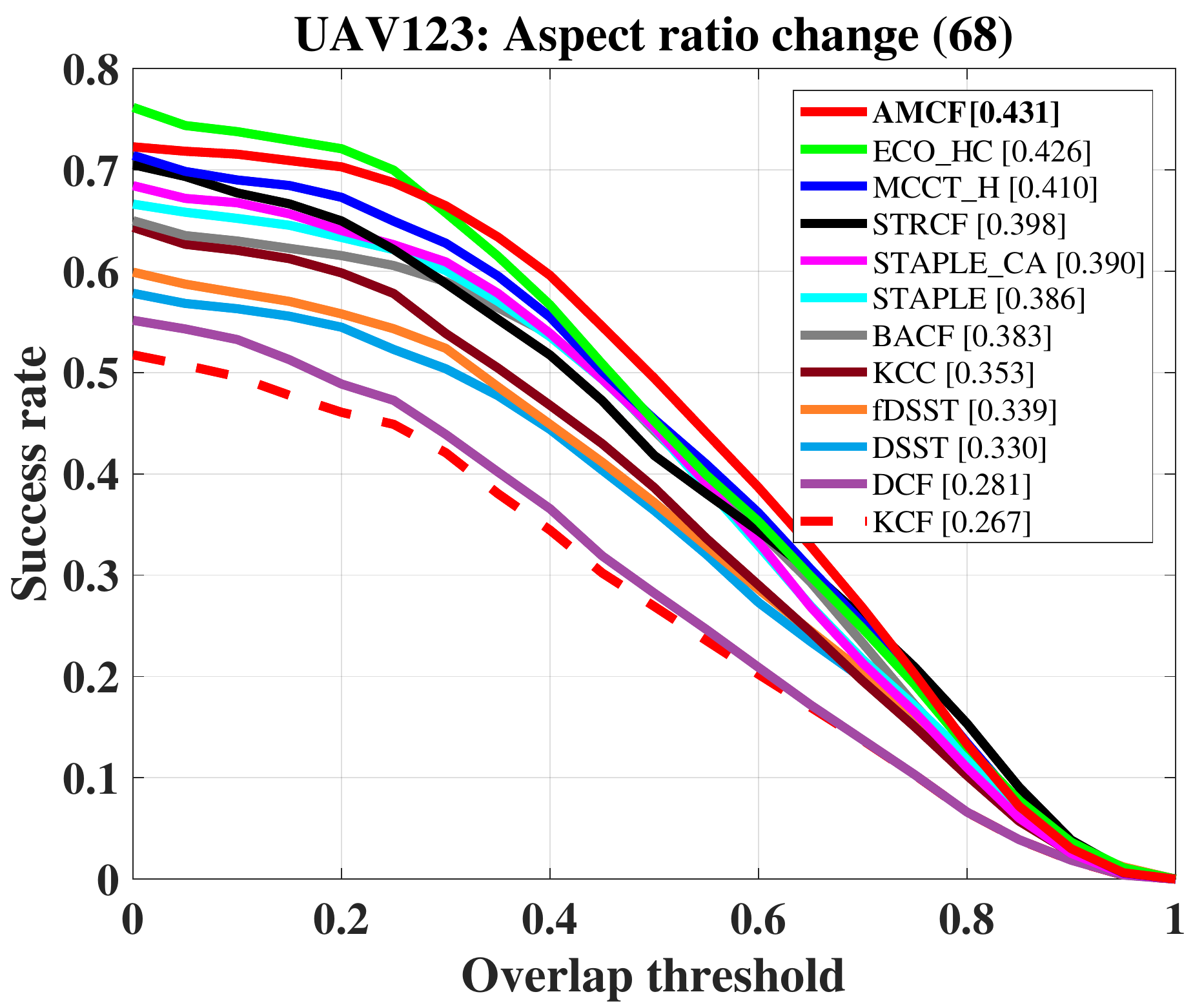}
	\caption{\textbf{Attribute-based evaluation.} Success plots of four attributes. The first two attributes are from UAVDT and the rest of them are from UAV123. }
	\label{fig:attribute}
\end{figure*}
\subsubsection{Attribute-based performance}
\label{subsec:attr}
Attribute-based evaluation results on both benchmarks are shown in Fig.~\ref{fig:attribute}. It can be seen that the reason behind our satisfactory generalization ability originates from the combination of two of our core modules, i.e., augmented memory and compressed context learning. On UAVDT, when object is small, feature of objects decreases and positive samples are thus not enough for robust tracking. Compressed context learning simultaneously expands search region and brings more negative samples. Therefore, when there is camera motion and background clutter, AMCF performs satisfactorily. On UAV123, object is significantly closer, so viewpoint change and aspect ratio change can result in more drastic appearance changes. Augmented memory provides more appearance information on previous objects so that a desired response can be obtained when current view has some resemblance to previous views. Therefore, thanks to both modules, ARCF can handle both near objects with large viewpoint changes and distant objects with a small size on the image.
\begin{figure}[!t]
	\centering
	\includegraphics[width=0.93\columnwidth]{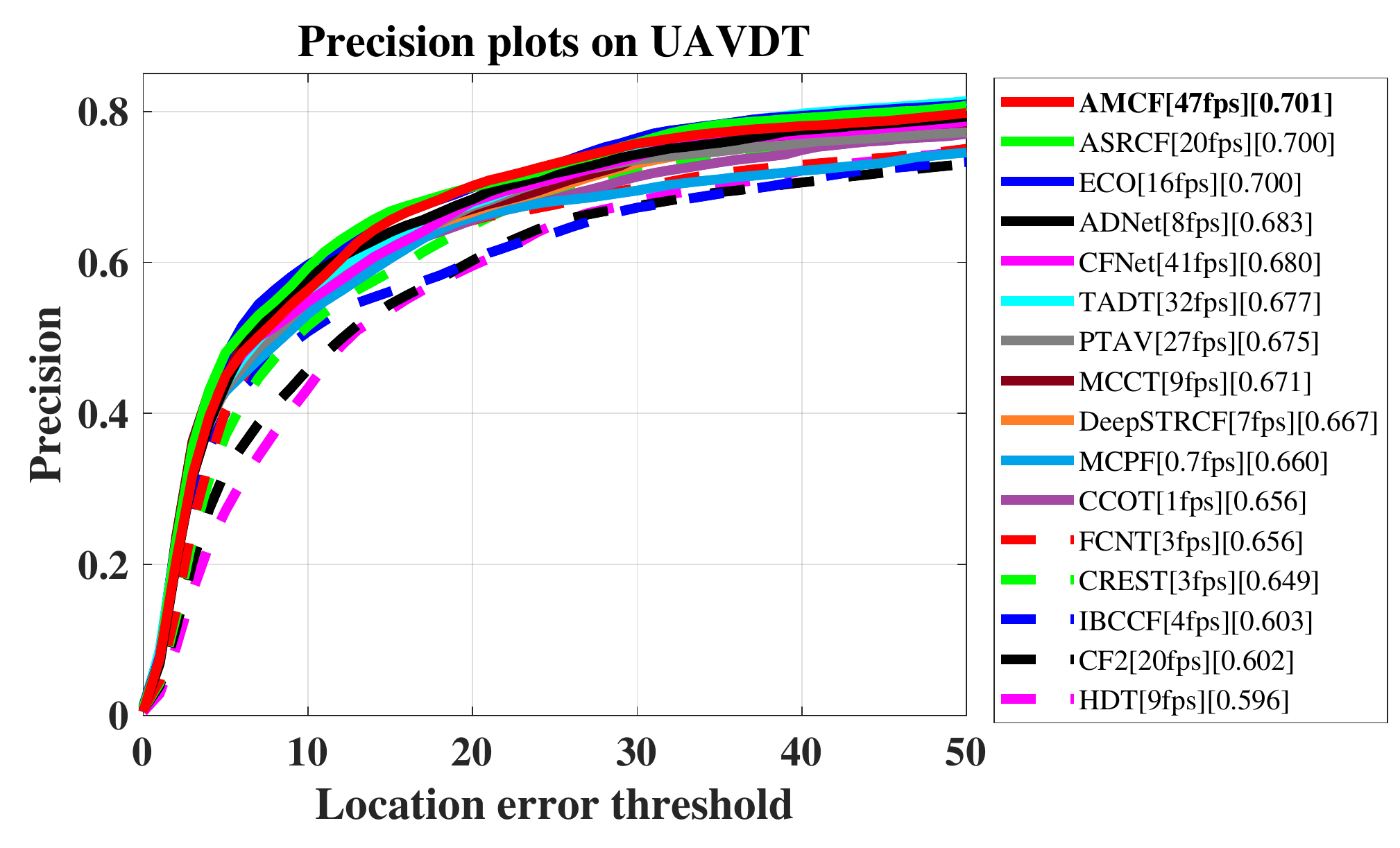}
	\\
	\includegraphics[width=0.93\columnwidth]{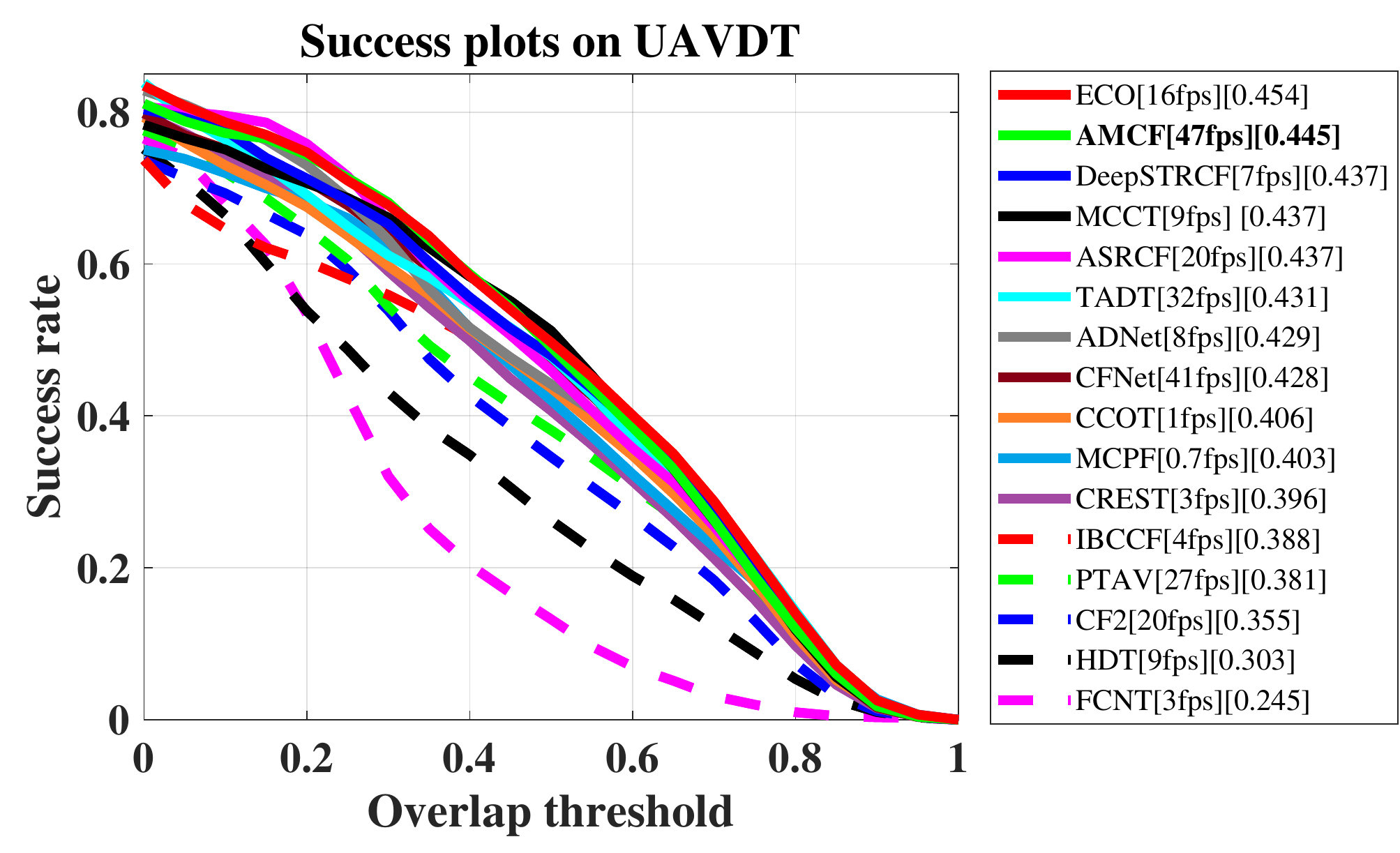}
	\caption{\textbf{Deep-based tracker comparison.} Top deep-based trackers are compared with AMCF on UAVDT. Tracking speeds other than AMCF are obtained on GPU.}
	\label{fig:deepcomparision}
\end{figure}
\subsection{Comparison with deep-based trackers}
On UAVDT, extra 15 deep-based trackers (trackers using deep learning methods or DCF-based trackers using deep features) are compared with AMCF. Precision and success plots can be seen in Fig.~\ref{fig:deepcomparision}. Surprisingly, AMCF still succeeds to exceed most of deep-based trackers in terms of precision and AUC scores. More specifically, AMCF achieved the best performance in precision, with 0.1\% slightly better than the second place ASRCF, while in AUC evaluation, AMCF achieved the second, falling behind ECO by only 0.9\%. In terms of tracking speed, AMCF is the fastest among the evaluated deep-based trackers. To sum up, in tracking distant objects, AMCF demonstrates superior tracking performance against both real-time trackers and deep-based trackers. 

\begin{figure}[t]
	\centering
	\includegraphics[width=0.95\columnwidth]{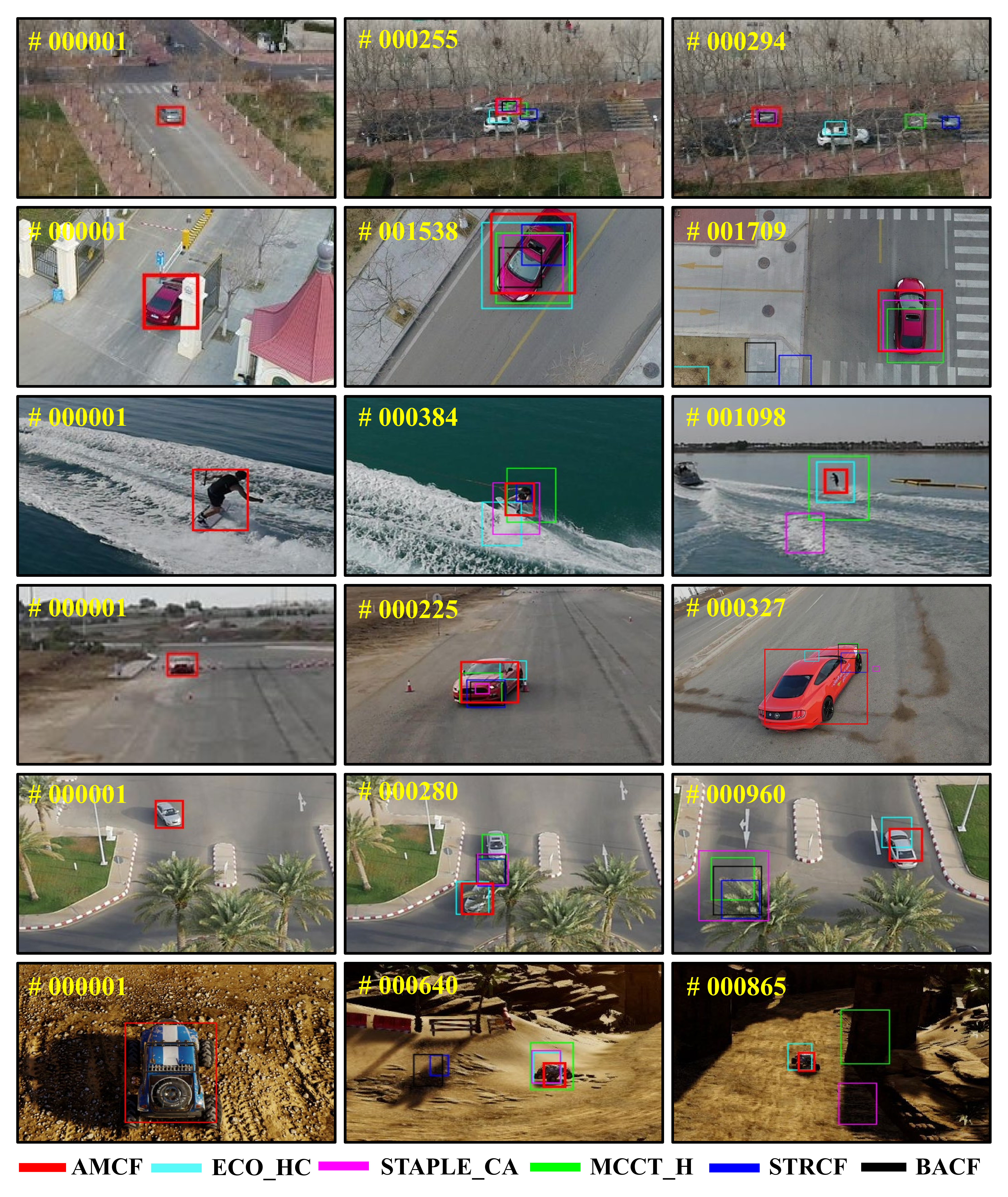}
	\caption{\textbf{Qualitative evaluation.} Top real-time trackers are compared with AMCF on \textit{S1001} and \textit{S1201} on UAVDT, as well as \textit{wakeboard5}, \textit{car16\_1}, \textit{car7}, and \textit{car1\_s} on UAV123. }
	\label{fig:comp}
\end{figure}
\subsection{Qualitative study}
Figure~\ref{fig:comp} intuitively demonstrates the aforementioned results. The first two sequences show the ability to track distant objects and adapt to view changes on UAVDT respectively. The third and fourth sequences show those abilities of AMCF on UAV123. Capability of resisting occlusion is demonstrated in the fifth sequence. The last one shows two modules can work smoothly together. 

\section{CONCLUSIONS}\label{sec:CONCLUSIONS}
In this work, augmented memory for correlation filters is proposed. Essentially, augmented memory maintains a FIFO queue to store distinct previous views. Each stored view in the memory will be assigned a desired response. Along with the feature of the object in the current frame, all views are simultaneously used to train a correlation filter that can adapt to new appearances and has response to previous views at the same time. Compressed context learning provides more negative samples and suppresses responses to surrounding areas of the tracked object. Extensive experiment results proved that AMCF has competitive performance and tracking speed compared to top real-time trackers. AMCF also demonstrates an outstanding generalization power that can track both near objects with large view change and distant objects with small size in the image. Future work can include introducing a confidence check to prevent false tracking results to be selected as a view. This method can also be applied to more powerful baselines in replacement of model update.


\section*{ACKNOWLEDGMENT}
This work is supported by the National Natural Science Foundation of China (No. 61806148) and the Fundamental Research Funds for the Central Universities (No. 22120180009).


\bibliographystyle{IEEEtran}  
\bibliography{IEEEabrv,ref}

\end{document}